\documentclass[journal]{IEEEtran}
\usepackage{graphicx}
\usepackage{amsmath}
\usepackage{acronym}
\usepackage{cite}
\usepackage{balance}

\newcommand{\matr}[1]{\begin{bmatrix}#1\end{bmatrix}}

\newcommand{\X}{{X}}
\newcommand{\Y}{{Y}}
\newcommand{\Z}{{Z}}

\newacro{CRT}{cathode ray tube}
\newacro{LUT}{look up table}
\newacro{LCD}{liquid crystal display}
\newacro{PCA}{principal component analysis}

\newacro{OLED}{organic light emitting diode}
\newacro{AMOLED}{active matrix organic light emitting diode}
\newacro{CIE}{Commission internationale de l'\'eclairage}

\begin{document}
\title{A statistical model of tristimulus measurements within and between OLED displays}
\author{Matti Raitoharju, Samu Kallio, and Matti Pellikka
\thanks{M. Raitoharju was with the 
Department of Automation Science and Engineering, Tampere University of Technology, Tampere, Finland, e-mail: matti.raitoharju@tut.fi, S. Kallio and M. Pellikka were with Microsoft Mobile Oy.}}

\maketitle

\begin{abstract}
We present an empirical model for noises in color measurements from OLED displays. According to measured data the noise is not isotropic in the XYZ space, instead most of the noise is along an axis that is parallel to a vector from origin to measured XYZ vector. The presented empirical model is simple and depends only on the measured XYZ values. Our tests  show that the variations between multiple panels of the same type  have similar distribution as the temporal noise in measurements from a single panel, but a larger magnitude.  
\end{abstract}


\section{Introduction}

\IEEEPARstart{T}{ristimulus} measurements from displays can be used for determining the color space of a display and calibration of the display. These measurements are commonly made in \ac{CIE} 1931 XYZ color space and can be used to determine the device independent color reproduction characteristics of displays. The components of XYZ color space are defined as integrals of product of a spectral radiance and color matching function over the visible light spectrum. The color matching functions for X, Y, and Z are positive. The use of XYZ color space ensures that no tristimulus value is negative \cite{Wyszecki}.
In practice, tristimulus measurements contain noise that is  partly caused by the spectrometer and partly due to the measured device. In \cite{badano2003principles}, several different sources for noise for \ac{CRT} and \ac{LCD} devices are presented.

In~\cite{JSID:JSID369} the angular optical properties of \ac{AMOLED}, \ac{OLED}-TV, and polymer \ac{OLED} displays are evaluated by making accurate measurements from various angles. In~\cite{boher2014spectral} the variation of displayed colors in different parts of a display is evaluated. 

Our work differs from earlier work in two main ways. First, we measure the each color multiple times from multiple panels. Our measurements provide insights to the variation of displayed colors in time and between panels, while related work on \ac{OLED} panels \cite{JSID:JSID369,boher2014spectral} is concentrated on measuring a single or a couple of displays from various locations and angles.

Second, we build a model that is based only on the statistical properties of the tristimulus measurements measured from \ac{OLED} displays. Using a physics-based approach as in \cite{badano2003principles} requires a better knowledge of the technical details of the measured panel, while the empirical model can be applied and fitted using only the measurements as the source of information and, thus, the model can be used and verified without access to the underlying hardware details. The statistical modeling takes the combination of all noise sources observed into account.

The measured XYZ values are usually converted into another color space, which is selected according to the application. The conversion from XYZ space to a RGB space, such as sRGB used in consumer devices, can be done using a linear transformation \cite{Colorimetry}. The XYZ values can also be converted to a uniform color space. The purpose of a uniform color space is to present the observed color differences \cite{mahy1994evaluation}. In calibration of the devices it is usually desired to minimize the observed color difference. The uncertainties defined in the XYZ measurement model can be transformed into other color spaces \cite{gardner2013tristimulus}. This allows the use of the proposed noise model also with other color spaces.

For color calibration there are several different models e.g.\ linear transformations \cite{806624}, \acp{LUT} \cite{806624}, polynomial models \cite{ccsmatlab}, and neural networks  \cite{ccsmatlab}. The color reproduction of \ac{OLED} panels is not linear. With uncalibrated panels in our test the luminance of pure white at maximum brightness was about 93\% of the luminance of the sum of luminances of the pure red, green, and blue. The proposed noise model for a single panel depends only on the measured XYZ values, which makes it independent of the colors sent to the panel and, thus, the calibration of the panel. 

In this paper, we build models for measurements from a single region of a single panel and between multiple panels. In measurements from a single panel the noise is temporal. In the measurements between panels there is, in addition to temporal noise,  variation that is caused by the different static properties of the different panels. We will show that even these noises are caused by different sources they have similar statistical properties. 

Rest of this paper is organized as follows. In Section~\ref{sec:model} we develop a noise model based on the empirical data. In Section~\ref{sec:ex} we show an example where the proposed model improves calibration. In Section~\ref{sec:comparison} we compare the amount of variation obtained from our measurements for single panel in single location at different times and between different panels with variation of different regions of an \ac{OLED} display presented in \cite{boher2014spectral}. Section~\ref{sec:conc} concludes the paper.

\section{Noise model for tristimulus measurements}
\label{sec:model}
\subsection{Background}
In tests with \ac{CRT} displays \cite{ohno1997four}, it is proposed that most of the noise is along the luminance component (Y).  In~\cite{gardner2013tristimulus} it is stated that the source noise in a spectral measurement contributes to uncertainties in each of the tristimulus values, but not to chromaticity values because of the correlations. The chromaticity values $x$ and $y$ are defined as
\begin{equation}
	x=\frac{X}{X+Y+Z}\text{ and }y=\frac{Y}{X+Y+Z}. \label{equ:chroma}
\end{equation}
If the noise does not affect $x$ and $y$ values, but affects $X$, $Y$, and $Z$ values, the noise vector has to be in the null space of Jacobian of the mapping \eqref{equ:chroma}. The null space in this case is aligned with $\matr{X&Y&Z}^T$ vector that makes the noise parallel with the $\matr{X&Y&Z}^T$ vector.

The assumption that the noise is aligned with $\matr{X&Y&Z}^T$ can be justified by looking at the definition of the color space components 
\begin{align}
	X&=\int L(\lambda ) \bar{x}(\lambda) \mathrm{d} \lambda \\
	Y&=\int L(\lambda ) \bar{y}(\lambda) \mathrm{d} \lambda \\
	Z&=\int L(\lambda ) \bar{z}(\lambda) \mathrm{d} \lambda,
\end{align}
where $L(\lambda)$ is the spectral radiance at wavelength $\lambda$, functions with bar are positive color matching functions, and the integral is made over the spectrum of visible light. If the spectral radiance is multiplied by a constant, then the direction of $\matr{X&Y&Z}^T$ does not change.  In next subsections, we test how the measured variation from displays vary along different directions to see if the largest variation is along $\matr{X&Y&Z}^T$ and what is its relation to noise along other vectors.

\subsection{Noise of measurements from an \ac{OLED} display}
\begin{figure*}[tb]
\centering
	\includegraphics[width=0.9\textwidth,clip=true,trim=0cm 0.5cm 0cm 0.5cm]{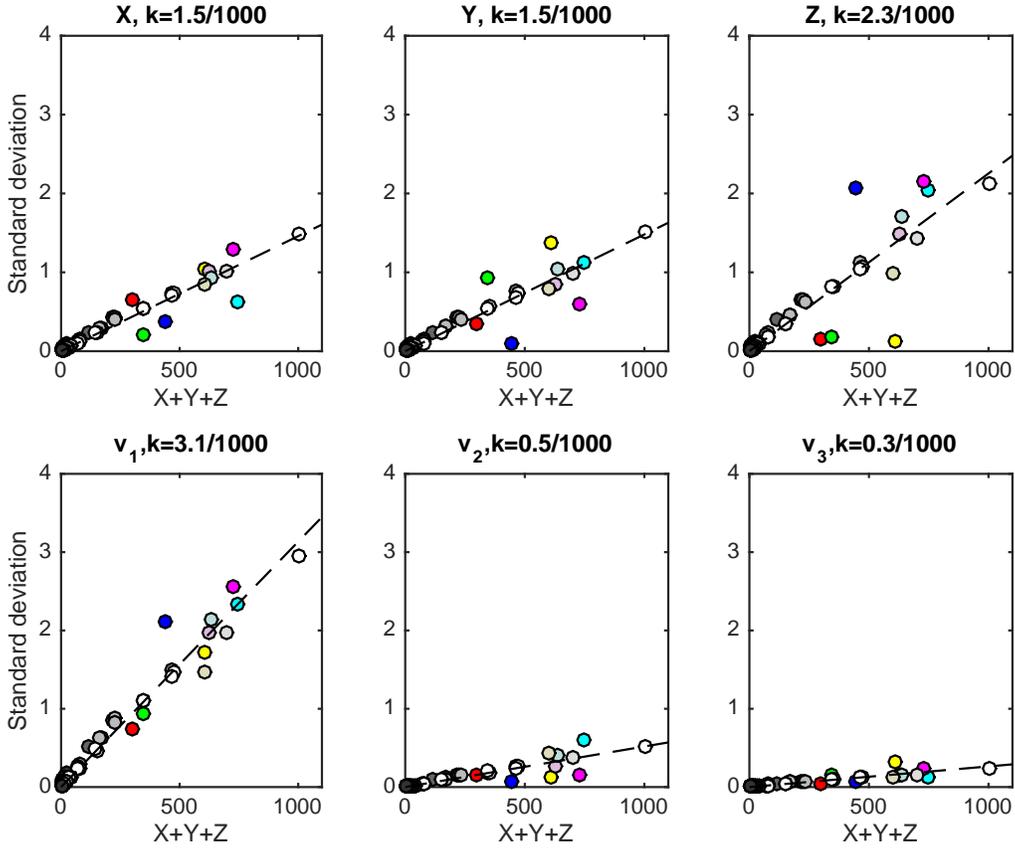}
		\caption{Standard deviation of measurements of a display along $X,Y,Z,v_1,v_2$, and $v_3$ of one panel}
	\label{fig:stddirections}
\end{figure*}
For evaluating the amount of noise in tristimulus measurements from an \ac{OLED} display, we measure several colors from a display, each color 12 times. The measurements are made with 4 calibrated Ocean Optics USB2000+ spectrometers (3 measurements with each spectrometer). The \ac{OLED} panel in this test is a 5.7" \ac{AMOLED} panel.

To compute the standard deviation along a direction we first compute the sample mean
\begin{equation}
	\mu = \frac{\sum_{i=1}^n c_i}{m},
\end{equation}
where $c_i$ is a tristimulus measurement and $m$ is the number of samples. The standard deviation along a unit vector $v$ is
\begin{equation}
	\sigma_{v} =   \sqrt{\sum_{i=1}^m \frac{ \left(v^T(c_i - \mu)\right)^2 }{m}} 
\end{equation}

Figure~\ref{fig:stddirections} shows the standard deviations along the main axis of the XYZ color-space, and along the $v_1=\frac{\matr{\X&\Y&\Z} }{\sqrt{\X^2+ \Y^2+\Z^2}}$ vector and two directions perpendicular to $v_1$ for a panel.  Each color was measured 12 times. 
The unit vectors are selected as follows:
\begin{align}
v_1 & = \frac{\matr{\X&\Y&\Z} }{\sqrt{\X^2+ \Y^2+\Z^2}} \\ v_2 & = \frac{\matr{0 & -\Z & \Y}^T}{ \sqrt{\Y^2+\Z^2}} \\ v_3 & = \frac{v_1 \times v_2}{\left\| v_1\times v_2 \right\|}.
\end{align}
Markers in the figure show the standard deviations for different colors, and the marker color presents the color drawn on the display. Some greys were measured in multiple brightness levels. Furthermore, the figure shows a linear fit that approximates
the standard deviations as a function of $X+Y+Z$:
\begin{equation}
	\sigma_j \approx k(X_j + Y_j +Z_j).
\end{equation}
For this problem, the linear least squares estimate for $k$ is
\begin{equation}
	k = \frac{\sum_{j=1}^n (X_j + Y_j +Z_j) \sigma_j}{\sum_{j=1}^n (X_j + Y_j +Z_j)^2},
\end{equation}
where $n$ is the number measured standard deviations.

The standard deviation along  $v_1$ is clearly larger than along the other dimensions. To see if this is a general behavior we repeat the same test for 12 other panels. Each color was measured with panels from 9 to 15 times. Table~\ref{tbl:dev} shows the values of $k \times 1000$ for the standard deviation models, the average $\mu$ of $k \times 1000$ of all panels and the standard deviation $\sigma$ of these. The standard deviation along $v_1$ is larger than the standard deviation along axis normal to it although there is some variation in the factor. This is caused by differences in displays and the relatively small sample size.   To verify the accuracy of the choice of $\matr{\X&\Y&\Z}^T$ direction we compute the axis of maximum variation using \ac{PCA}. The  $\matr{\X&\Y&\Z}^T$ and the main axis of the noise have a difference $14$ degrees on average. In comparison, assuming the main axis has a 60-degree angle with $Y$ axis on average. This shows that the assumption of having most of the noise along Y vector in \cite{ohno1997four}, does not hold for \ac{OLED} displays.

\setlength{\tabcolsep}{3pt}
\begin{table*}[tbh]
\caption{Factor of standard deviations $\times 1000$ of tristimulus measurements along and normal to $[\X,\Y,\Z]^T$ measured from 5.7" \ac{AMOLED} panel} \label{tbl:dev}
\small
\centering
\begin{tabular}{c|ccccccccccccc|cc}
display & 1& 2& 3& 4& 5& 6& 7& 8& 9& 10& 11& 12& 13  & $\mu$ & $\sigma$ \\ \hline
$v_1$ & 3.1 & 1.4 & 1.8 & 1.7 & 2.2 & 2.8 & 2.1 & 1.8 & 2.4 & 2.0 & 1.1 & 1.4 & 2.9 &  2.1 & 0.6\\
$v_2$& 0.5 & 0.8 & 1.0 & 0.8 & 0.7 & 0.8 & 0.6 & 0.6 & 1.0 & 0.5 & 0.7 & 0.9 & 1.1 & 0.8 & 0.2 \\
$v_3$ &  0.3 & 0.4 & 0.4 & 0.3 & 0.3 & 0.4 & 0.3 & 0.2 & 0.3 & 0.3 & 0.2 & 0.3 & 0.6 & 0.3 &0.1\\
\end{tabular}
\end{table*}
The standard deviation is in every sample larger along vector $v_2$ than along $v_3$. With this data, the factor of standard deviation along $[\X,\Y,\Z]$ compared to a normal direction is on average~3.7.  Because this test was made with 4 different spectrometers, even though they all were of similar type and calibrated, there may be some calibration related noise in the data.

To verify the results, we tested a different \ac{AMOLED} panel with 5.2" diameter. In this test, we measured white at maximum brightness setting from each panel 10 times and computed the standard deviations for the measurements. These measurements were done with a single spectrometer within 15 minutes' time. Table~\ref{tbl:devB} shows the standard deviations of measurements of the white.
 \begin{table*}[tpb]
\caption{Factor of standard deviations $\times 1000$ of tristimulus measurements along and normal to $[\X,\Y,\Z]^T$ when measured from a 5.2" \ac{AMOLED} panel} \label{tbl:devB}
\small
\centering
\begin{tabular}{c|ccccccccccc|cc}
display & 1& 2& 3& 4& 5& 6& 7& 8& 9& 10& 11 & $\mu$ & $\sigma$ \\ \hline
$v_1$ & 3.8 & 2.8 & 1.8 & 2.3 & 1.7 & 2.8 & 2.0 & 1.3 & 1.8 & 2.5 & 1.9 & 2.2  & 0.7 \\
$v_2$ &0.4 & 0.3 & 0.2 & 0.3 & 0.3 & 0.3 & 0.4 & 0.1 & 0.2 & 0.3 & 0.3 & 0.3 & 0.1 \\
$v_3$ &0.3 & 0.4 & 0.2 & 0.2 & 0.4 & 0.4 & 0.9 & 0.4 & 0.2 & 0.3 & 0.2 & 0.4 & 0.2  
\end{tabular}
\end{table*}
The results are similar as with the 5.7" display. We can note that the standard deviation along $v_1$ and other directions is larger at 6.9 and the variation along $v_3$ is larger than along $v_2$, which is opposite than with the results in Table~\ref{tbl:dev}.

\begin{figure}[tb]
	\centering
	\includegraphics[width=\columnwidth]{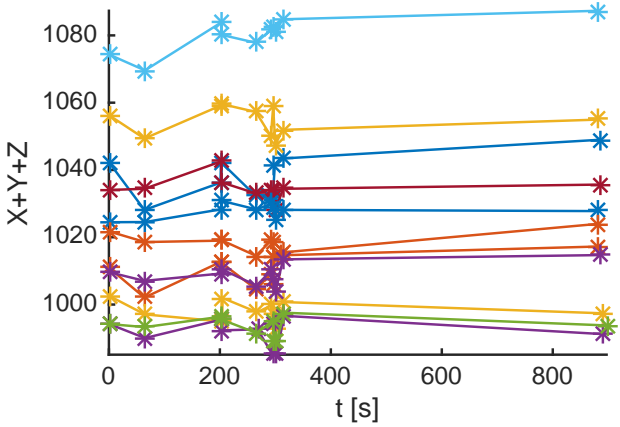}
	\caption{Value of X+Y+Z measured from 11 \ac{OLED} panels showing white color at different time instances. Each line corresponds to a different panel and asterisk to a measurement time.}
	\label{fig:Bintime}
\end{figure}

Figure~\ref{fig:Bintime} shows the value of $X+Y+Z$ of the white as the function of the measurement time from each display. Asterisks show the measurements and each color marks a panel. This figure shows  no clear trends in measurement values during 15 minute measurement sequence, which could have been caused be warming of the panel in use, and that measurements from each panel had approximately similar amount of noise and the variations between panels are larger than variations within panel. 

\subsection{Noise of measurements between \ac{OLED} displays}
Next we test how the standard deviation of the tristimulus measurements varies between panels. For 5.7" panel used test sequence that contains 765 measurements in total. The measurements were done using 3 different brightness levels i.e.\ there are 255 values colors given to panel so that the panel brightness is set to 100\%, 50\% and 25\%. This measurement sequence was done for 22 panels.  For 5.2" panels we used two test sequences. A shorter test sequence of 50 measurements was used for 702 panels and longer with 765 measurements to 11 panels. Each color was measured only once from a panel.

\begin{figure*}[tb]
	\centering
	\includegraphics[width=0.9\textwidth,clip=true,trim=0cm 1.5cm 0cm 0.5cm]{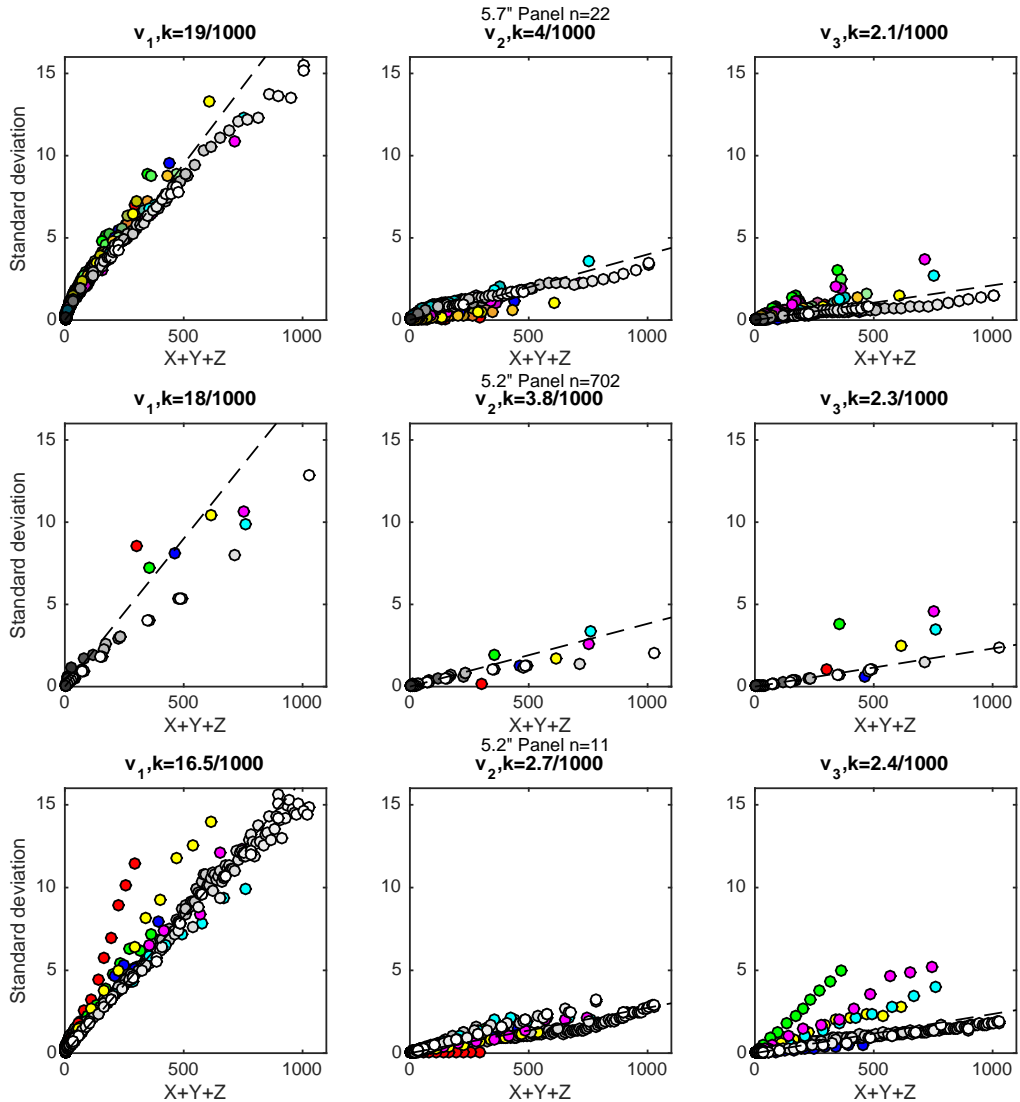}
	\caption{Standard deviation of measurements along $[\X,\Y,\Z]^T$ ($v_1$) and 2 perpendicular directions ($v_1,v_2$) of multiple panels}
	\label{fig:alldisplays}
\end{figure*}

Figure~\ref{fig:alldisplays} shows the standard deviations data. For all test sequences the standard deviation along $v_1$ is larger than along the two other dimensions. 
The result is similar to the result achieved by measuring only one display, but the amount of standard deviation is approximately sixfold compared to variations within display.

Figures also show that between panels there is more variation in some colors. Specifically, green has relatively large variation along $v_3$, although it is still smaller than the variation along $v_1$. 

\subsection{Noise model}
We construct a statistical model for noise from \ac{OLED} displays based on the experiments in the two previous subsections. The covariance matrix uses 5-fold measurement noise along $\matr{\X & \Y &\Z}^T$ compared to other directions. It can be written as
\begin{equation}
P=a^2 (X+Y+Z)^2 U\matr{ 5^2 & 0 & 0 \\ 0 & 1 & 0 \\ 0 & 0 &1} U^T, \label{equ:P}
\end{equation}
where
\begin{equation}
U  =\matr{v_1 & v_2 & v_3}
\end{equation}
and $a$ is a multiplier that can be set to $\frac{1}{2000}$ for measurements within display and $\frac{1}{400}$ for measurements between displays. If we assume that the temporal noise in every panel same and the differences between panels does not change in time, after an optimal calibration the factor $a$ for perfectly calibrated panels would be $\frac{1}{2000}$. On the other hand, the temporal noise of measurements can be reduced by making multiple measurements and taking an average of them. Taking average of multiple measurements increases the measurement time.

In many applications, the value of $a$ does not affect the result and can be omitted. In the next section, we show one such application.

\section{Use of noise model in calibration}
\label{sec:ex}
In this application, we try to find a matrix $M$ such that measurement  $c_i=[\X_i, \Y_i, \Z_i]^T$ from a display multiplied with it matches with a measurement $\hat c_i$ from another display. That is
\begin{equation}
	\hat c_i \approx M c_i. \label{equ:CT}
\end{equation}

To fit multiple measurements, we minimize the norm
\begin{equation}
	\sum_i \left( \hat c_i - M c_i   \right)^T\left(\hat P_i + MP_iM^T\right)^{-1}\left( \hat c_i - M c_i   \right),
\end{equation}
where $P_i$ is the covariance (\ref{equ:P}) for measurement $c_i$ and $\hat P_i$ is covariance for $\hat c_i$.

There is no simple solution for this equation, because $M$ term is used to induce the norm and also transforming the measurement. 
Because the target is to transform measurements $c_i$ to match the reference measurements we can make approximation 
\begin{equation}
   MP_iM^T  \approx  \hat P_i 
\end{equation}
Now the minimized function becomes
\begin{equation}
\sum_i \left( \hat c_i - M c_i \right)^T (2 \hat P_i)^{-1}\left( \hat c_i - M c_i \right).
\end{equation}
One measurement can be written
\begin{equation}
	\overbrace{\matr{\hat X_i\\\hat Y_i\\ \hat Z_i}}^{\hat c_i}  = \overbrace{\matr{X_i & Y_i& Z_i & 0 & 0 & 0 &0 &0 &0 \\
	 0 &0 &0 & X_i & Y_i& Z_i & 0 & 0 &0 \\
	0 & 0 & 0 &0 &0 &0  & X_i & Y_i& Z_i }}^{H_i} \overbrace{\matr{ m_{1,1} \\ m_{1,2} \\ m_{1,3} \\ m_{2,1} \\ m_{2,2} \\ m_{2,3} \\ m_{3,1} \\ m_{3,2} \\m_{3,3}}}^{m}  +\varepsilon_i,
\end{equation}
where $\varepsilon_i$ are independent zero-mean multivariate Gaussians with covariance $\hat{P}_i$, and $m_{i,j}$ is an element of $M$. 

To compute the optimal estimate, the weighted linear least squares formula is used
\begin{equation}
	m=\left(H^TWH\right)^{-1}H^TW\hat{c}, \label{equ:M}
\end{equation}
where
\begin{align}
	\hat{c}&=\matr{\hat c_1 \\ \vdots \\ \hat c_n} \\
	 H&=\matr{H_1 \\ \vdots \\ H_n} \\
	 W&=\mathrm{diag}\, ((2\hat{P}_1)^{-1}, \ldots, (2\hat{P}_n)^{-1}).
\end{align}
This formulation requires at least 3 linearly independent measurements, which can be obtained e.g.\ by measuring pure R, G, and B colors. When using only 3 measurements the weighting with noise model does not affect to the values of matrix $M$, but with more measurements it has an effect on the result. Also, we can note that $M$ is independent of the multiplier $a$.

This algorithm is similar to the algorithm used in \cite{gardner2013tristimulus} for colorimeter calibration, but here we calibrate outputs of two displays and use a variable covariance weighting for different dimensions in colorspace. The benefit over the four-color method proposed in \cite{ohno1997four} is that this formulation allows using any number ($\geq 3$)  of measurements. 

\begin{figure}
\centering
	\includegraphics[width=\columnwidth,clip=true,trim=0.8cm 0cm 1.35cm 0cm]{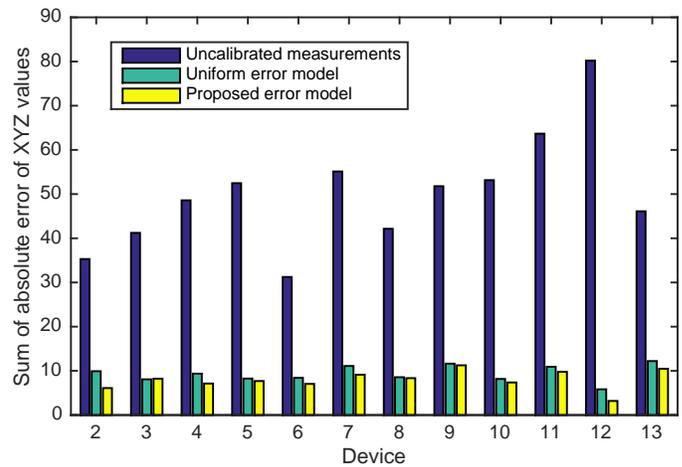}
	\caption{Results of the calibration using the proposed error model}
	\label{fig:displaycalibres}
\end{figure}

We test the proposed model in calibration of display outputs. We measure red, green, blue and white from a pair of mobile terminal displays and fit matrix $M$ using these measurements. Elements of $M$ are computed using (\ref{equ:M}). For verification of the $M$ we measure cyan, magenta, and yellow from the displays and transform obtained tristimulus measurements from the first display with the matrix (\ref{equ:CT}). The result of the transformation is compared with the measurements from the other display. Figure~\ref{fig:displaycalibres} shows the sum of absolute values of differences of the XYZ values when device 1 is compared with other 12 devices.  Uniform error model assumes that the error is distributed uniformly while the proposed model assumes most noise is along $v_1$.
Figure shows how the use of the proposed error model reduces the error compared to the uniform error model in 11 calibrations out of 12. The calibration accuracy is improved even the measurement set contained green color that was earlier found out to have larger variations than the model would predict. 

\section{Comparison to related work}
\label{sec:comparison}
In~\cite{boher2014spectral} different regions of a 6" \ac{AMOLED} display are measured so that the display shows one of primary colors at the time. Then the variation of the $\Delta E$ of the measured tristimulus values compared to the average is evaluated. The $\Delta E$ is thus 
\begin{equation}
\Delta E = \sqrt{ (L^*-L_\text{avg}^*)^2 + (a^*-a_\text{avg}^*)^2 + (b^*-b_\text{avg}^*)^2},
\end{equation}
where $L^*$, $a^*$, and $b^*$ are the color components of the \ac{CIE} 76 color system and variables with subindex avg are the average values from the whole display.

We compute the $\Delta E$ values of our measurements for variation when measurement is repeated multiple times at one region of the display for one panel and for variation of colors of multiple panels. The measurements for variation of one location in one display is made using the 12 5.7" displays. The average values are computed for each display separately. The variation of multiple panels is computed for 702 5.2" displays.

\begin{figure}[tb]
\centering
	\includegraphics[width=\columnwidth,clip=true,trim=0cm 1.2cm 1cm 0.5cm]{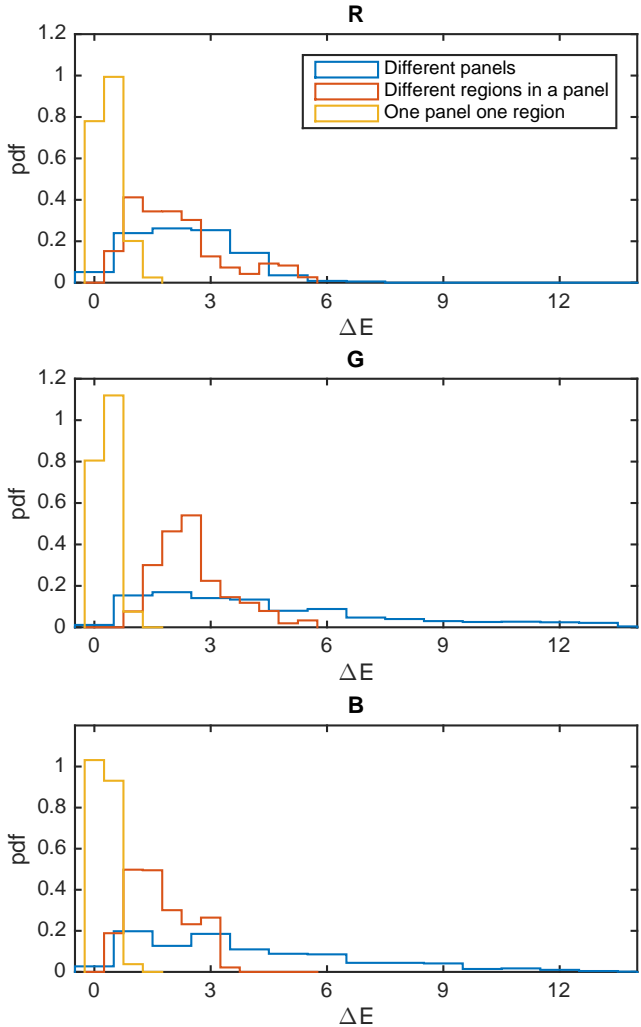}
	\caption{Histograms of $\Delta E$ obtained by measuring one region of one display multiple times, measuring different regions of one \ac{OLED} display \cite{boher2014spectral}, and measuring different \ac{OLED} displays of the same type}
	\label{fig:comparison}
\end{figure}

Figure~\ref{fig:comparison} shows the histograms of $\Delta E$ for red, green, and blue computed from our data, different panels and  one panel one region, and the data for different regions in a panel obtained from Fig.~4 of~\cite{boher2014spectral}.
The histograms show that our measurements are consistent with the results presented in \cite{boher2014spectral}. Measurements from one panel at one region have smaller variation than when measuring from multiple regions and variation between different panels is larger than variation within one panel. The mean of $\Delta E$ between panels is 3.6, between regions of a panel $1.8$ and in one region in one panel $0.3$. Just noticeable difference of $\Delta E$ is about 2.3 \cite{COL:COL70}, which means that the variations of one display in one region are not detectable by human eye. The variation between different parts of a display may be just noticeable. Different displays of same type may have color differences that are well noticeable. 

\section{Conclusions and future work}
\label{sec:conc}
\balance

In this paper, we developed an empirical model for tristimulus measurements from \ac{OLED} displays. In our model, most of variation is aligned along $\matr{X & Y & Z}^T$ vector and the measurements from \ac{OLED} displays support this. In~\cite{gardner2013tristimulus} was proposed that all of the variation is along this direction. We analyzed the variations and did find there is variation in directions perpendicular to $\matr{X & Y & Z}^T$. We found out that the variations are almost linearly dependent on the sum $X + Y + Z$ and the ratio between variation along $\matr{X & Y & Z}^T$ and directions perpendicular to it is approximately 5.

The variation of one display at one region is temporal in its nature, while most of the variation between displays is static so it cannot be explained by temporal type of noises, such as shot noise. However, results showed that the variation of tristimulus values measured from one display at different time instances is similar to variation between displays, but with a different scale.  We compared to our results of amount of variation with variation between different regions of a display presented in \cite{boher2014spectral}. Expectedly the variation measured from one display and one region was smaller and variation between displays was larger than the variation between regions of one display. In calibration, the within panel model can be interpreted to be the lower bound of calibration accuracy that can be achieved. However, it may be unfeasible to make the calibration separately for different parts of a display and thus the variation in different regions of a display is the lower bound in practice.

The proposed model is fitted using measurements so it covers all sources of noise that contributed to our measurement. \acp{OLED} are current driven \cite{4068955} and the source-drain current uniformity  is a major factor in the short-term variations in \ac{OLED} displays \cite{SDTP:SDTP3976}. In \cite{JSID:JSID879} (as cited in \cite{boher2014spectral}) variations in subpixel level radiance was determined to be caused by current fluctuations.  In \cite{4068955} it is stated that process variation in the electrical characteristics, such as threshold voltage, results in differences in the OLED current among pixels. Based on these references, we hypothesize that the variations are mostly caused by current. The variation of color, when measured from one location of a panel would be caused the fluctuation of current in time and the variation between displays would be caused variation of static properties of the electrical characteristics of the displays. 

The proposed model is simple, does not require a deep knowledge of technical properties of the panel, and can be fitted using a relatively small sample size (models in Fig.~\ref{fig:alldisplays} for 5.2" panels were roughly the same with $n=11$ and $n=702$). The variations of some colors e.g.\ green may differ from the model, but our test results showed that still the proposed model improves accuracy over the uniform error model.

The developed model was tested in calibration of tristimulus measurements from different displays. Results showed that the proposed model achieved better accuracy than a uniform error model. The calibration test converted the measured XYZ values from a panel to expected values from another panel. Due to the nonlinear nature of the \ac{OLED} displays the linear calibration is not directly suitable for calibrating the input of the panel and the output, but we expect that using the proposed model with other calibration models \cite{806624,ccsmatlab} would improve the calibration accuracy.

The measurements were made with two different types of panels, both from a same manufacturer. In future, the model should be verified with different types of \ac{OLED} panels and panels that use different display technology. Because the model is simple and based only on measurements it is straightforward to build this kind of model and see if the other panel types exhibit similar properties. 

\bibliographystyle{IEEEtran}
\bibliography{viitteetspectrometer}{}

\begin{thebibliography}{10}
\providecommand{\url}[1]{#1}
\csname url@samestyle\endcsname
\providecommand{\newblock}{\relax}
\providecommand{\bibinfo}[2]{#2}
\providecommand{\BIBentrySTDinterwordspacing}{\spaceskip=0pt\relax}
\providecommand{\BIBentryALTinterwordstretchfactor}{4}
\providecommand{\BIBentryALTinterwordspacing}{\spaceskip=\fontdimen2\font plus
\BIBentryALTinterwordstretchfactor\fontdimen3\font minus
  \fontdimen4\font\relax}
\providecommand{\BIBforeignlanguage}[2]{{%
\expandafter\ifx\csname l@#1\endcsname\relax
\typeout{** WARNING: IEEEtran.bst: No hyphenation pattern has been}%
\typeout{** loaded for the language `#1'. Using the pattern for}%
\typeout{** the default language instead.}%
\else
\language=\csname l@#1\endcsname
\fi
#2}}
\providecommand{\BIBdecl}{\relax}
\BIBdecl

\bibitem{Wyszecki}
G.~Wyszecki and W.~Stiles, \emph{Color Science Concepts and Methods,
  Quantitative Data and Formulae}, 2nd~ed.\hskip 1em plus 0.5em minus
  0.4em\relax John Wiley and Sons, 1982.

\bibitem{badano2003principles}
A.~Badano, ``Principles of cathode-ray tube and liquid crystal display
  devices,'' \emph{Advances in digital radiography: {RSNA} categorical course
  in diagnostic radiology physics}, pp. 91--102, 2003.

\bibitem{JSID:JSID369}
\BIBentryALTinterwordspacing
P.~Boher, T.~Leroux, V.~Collomb-Patton, and T.~Bignon, ``Optical
  characterization of {OLED} displays,'' \emph{Journal of the Society for
  Information Display}, vol.~23, no.~9, pp. 429--437, 2015. [Online].
  Available: \url{http://dx.doi.org/10.1002/jsid.369}
\BIBentrySTDinterwordspacing

\bibitem{boher2014spectral}
P.~Boher, T.~Leroux, T.~Bignon, and V.~C. Patton, ``Spectral imaging analysis
  of oled display light emission properties,'' \emph{Proc. 21st Int'l Disp.
  Worksh}, 2014.

\bibitem{Colorimetry}
N.~Ohta and A.~R. Robertson, \emph{Colorimetry : fundamentals and
  applications}, ser. Wiley-IS\&T series in imaging science and
  technology.\hskip 1em plus 0.5em minus 0.4em\relax Chichester, GB, Hoboken,
  NJ, USA: John Wiley and Sons, 2005.

\bibitem{mahy1994evaluation}
M.~Mahy, L.~Eycken, and A.~Oosterlinck, ``Evaluation of uniform color spaces
  developed after the adoption of {CIELAB} and {CIELUV},'' \emph{Color Research
  \& Application}, vol.~19, no.~2, pp. 105--121, 1994.

\bibitem{gardner2013tristimulus}
J.~Gardner, ``Tristimulus colorimeter calibration matrix uncertainties,''
  \emph{Color Research \& Application}, vol.~38, no.~4, pp. 251--258, 2013.

\bibitem{806624}
M.~Vrhel and H.~Trussell, ``Color device calibration: a mathematical
  formulation,'' \emph{Image Processing, IEEE Transactions on}, vol.~8, no.~12,
  pp. 1796--1806, Dec 1999.

\bibitem{ccsmatlab}
S.~Westland, C.~Ripamonti, and C.~Vien, \emph{Computational Colour Science
  Using {M}atlab}, 2nd~ed.\hskip 1em plus 0.5em minus 0.4em\relax Wiley, 2004.

\bibitem{ohno1997four}
Y.~Ohno and J.~E. Hardis, ``Four-color matrix method for correction of
  tristimulus colorimeters,'' in \emph{Color and Imaging Conference}, vol.
  1997.\hskip 1em plus 0.5em minus 0.4em\relax Society for Imaging Science and
  Technology, 1997, pp. 301--305.

\bibitem{COL:COL70}
\BIBentryALTinterwordspacing
M.~Mahy, L.~Van~Eycken, and A.~Oosterlinck, ``Evaluation of uniform color
  spaces developed after the adoption of cielab and cieluv,'' \emph{Color
  Research \& Application}, vol.~19, no.~2, pp. 105--121, 1994. [Online].
  Available: \url{http://dx.doi.org/10.1111/j.1520-6378.1994.tb00070.x}
\BIBentrySTDinterwordspacing

\bibitem{4068955}
C.~L. Lin and Y.~C. Chen, ``A novel {LTPS-TFT} pixel circuit compensating for
  {TFT} threshold-voltage shift and {OLED} degradation for {AMOLED},''
  \emph{IEEE Electron Device Letters}, vol.~28, no.~2, pp. 129--131, Feb 2007.

\bibitem{SDTP:SDTP3976}
\BIBentryALTinterwordspacing
T.~Tsujimura, Y.~Kobayashi, K.~Murayama, A.~Tanaka, M.~Morooka, E.~Fukumoto,
  H.~Fujimoto, J.~Sekine, K.~Kanoh, K.~Takeda, K.~Miwa, M.~Asano, N.~Ikeda,
  S.~Kohara, S.~Ono, C.-T. Chung, R.-M. Chen, J.-W. Chung, C.-W. Huang, H.-R.
  Guo, C.-C. Yang, C.-C. Hsu, H.-J. Huang, W.~Riess, H.~Riel, S.~Karg,
  T.~Beierlein, D.~Gundlach, S.~Alvarado, C.~Rost, P.~Mueller, F.~Libsch,
  M.~Mastro, R.~Polastre, A.~Lien, J.~Sanford, and R.~Kaufman, ``4.1: A 20-inch
  {OLED} display driven by super-amorphous-silicon technology,'' \emph{SID
  Symposium Digest of Technical Papers}, vol.~34, no.~1, pp. 6--9, 2003.
  [Online]. Available: \url{http://dx.doi.org/10.1889/1.1832193}
\BIBentrySTDinterwordspacing

\bibitem{JSID:JSID879}
\BIBentryALTinterwordspacing
D.-Y. Shin, J.-K. Woo, Y.~Hong, K.-N. Kim, D.-I. Kim, M.-H. Yoo, H.-D. Kim, and
  S.~Kim, ``Reducing image sticking in amoled displays with time-ratio gray
  scale by analog calibration,'' \emph{Journal of the Society for Information
  Display}, vol.~17, no.~9, pp. 705--713, 2009. [Online]. Available:
  \url{http://dx.doi.org/10.1889/JSID17.9.705}
\BIBentrySTDinterwordspacing

\end{thebibliography}

\end{document}